\documentclass[11pt,a4paper]{article}
  \pdfoutput=1
  \RequirePackage[hyphens]{url}
  \usepackage[hyperref]{acl2018}
  \usepackage{times}
  \usepackage{latexsym}

  \usepackage{booktabs}
  \usepackage{listings}

  \usepackage{graphicx}
  \usepackage{tikz}
  \usepackage{pgfplots}
  \usepackage{pgfplotstable}
  \usepgfplotslibrary{groupplots}
  \pgfplotsset{compat=newest}
  \usepackage{siunitx}
  \usepackage{amsmath,amssymb,bm,bbm}
  \usepackage{datatool}
  \usepackage{etoolbox}
  \usepackage{stfloats}
  \usepackage{amsmath,amssymb,bm,bbm}
  \usepackage{datatool}
  \usepackage{etoolbox}
  \usepackage{stfloats}
  \usepackage{filecontents}
  \usepackage{enumitem}
  \setlist{noitemsep}

  \definecolor{bblue}{HTML}{4F81BD}
  \definecolor{rred}{HTML}{C0504D}
  \definecolor{ggreen}{HTML}{9BBB59}
  \definecolor{ppurple}{HTML}{9F4C7C}
  \definecolor{oorange}{HTML}{F08000}

  \aclfinalcopy 

  \title{Marian: Fast Neural Machine Translation in C++}

  \author{Marcin Junczys-Dowmunt$^{\dagger}$ \, Roman Grundkiewicz$^{*}$$^{\ddagger}$ \, Tomasz Dwojak$^{*}$ \\ {\bf Hieu Hoang \, Kenneth Heafield$^{\ddagger}$ \, Tom Neckermann$^{\ddagger}$ }  \\  {\bf Frank Seide$^{\dagger}$ \, Ulrich Germann$^{\ddagger}$ \, Alham Fikri Aji$^{\ddagger}$ } \\ { \bf Nikolay Bogoychev$^{\ddagger}$ \, Andr\'{e} F. T. Martins$^{\mathparagraph}$ \, Alexandra Birch$^{\ddagger}$} \\[2mm]
  $^{\dagger}$Microsoft \, $^{*}$Adam Mickiewicz University in Pozna\'{n} \\ $^{\ddagger}$University of Edinburgh \,  $^{\mathparagraph}$Unbabel }

  \date{}

\begin{document}
  \maketitle
  \begin{abstract}
  We present Marian, an efficient and self-contained Neural Machine Translation framework with an integrated automatic differentiation engine based on dynamic computation graphs. Marian is written entirely in C++. We describe the design of the encoder-decoder framework and demonstrate that a research-friendly toolkit can achieve high training and translation speed.  
  \end{abstract}

  \section{Introduction}

  In this paper, we present Marian,\footnote{Named after Marian Rejewski, a Polish mathematician and cryptologist who reconstructed the German military Enigma cipher machine sight-unseen in 1932. \url{https://en.wikipedia.org/wiki/Marian_Rejewski}.} an efficient Neural Machine Translation framework written in pure C++ with minimal dependencies. It has mainly been developed at the Adam Mickiewicz University in Pozna\'{n} and at the University of Edinburgh. It is currently being deployed in multiple European projects and is the main translation and training engine behind the neural MT launch at the World Intellectual Property Organization.\footnote{\url{https://slator.com/technology/neural-conquers-patent-translation-in-major-wipo-roll-out/}}

  In the evolving eco-system of open-source NMT toolkits, Marian occupies its own niche best characterized by two aspects:
  \begin{itemize}
  \item It is written completely in C++11 and intentionally does not provide Python bindings; model code and meta-algorithms are meant to be implemented in efficient C++ code.
  \item It is self-contained with its own back end, which provides reverse-mode automatic differentiation based on dynamic graphs.
  \end{itemize}

  Marian has minimal dependencies (only Boost and CUDA or a BLAS library) and enables barrier-free optimization at all levels: meta-algorithms such as MPI-based multi-node training, efficient batched beam search, compact implementations of new models, custom operators, and custom GPU kernels. Intel has contributed and is optimizing a CPU backend. 


  Marian grew out of a C++ re-implementation of Nematus \cite{sennrich-EtAl:2017:EACLDemo}, and still maintains binary-compatibility for common models. Hence, we will compare speed mostly against Nematus. OpenNMT \cite{klein2017opennmt}, perhaps one of the most popular toolkits, has been reported to have training speed competitive to Nematus. 

  Marian is distributed under the MIT license and available from \url{https://marian-nmt.github.io} or the GitHub repository \url{https://github.com/marian-nmt/marian}.

  \section{Design Outline}

  We will very briefly discuss the design of Marian. Technical details of the implementations will be provided in later work.

  \subsection{Custom Auto-Differentiation Engine}

  The deep-learning back-end included in Marian is based on reverse-mode auto-differentiation with dynamic computation graphs and among the established machine learning platforms most similar in design to DyNet \cite{dynet}. While the back-end could be used for other tasks than machine translation, we choose to optimize specifically for this and similar use cases. Optimization on this level include for instance efficient implementations of various fused RNN cells, attention mechanisms or an atomic layer-normalization \cite{ba2016layer} operator. 

  \subsection{Extensible Encoder-Decoder Framework}
  \label{encdec}

  Inspired by the stateful feature function framework in Moses \cite{conf/acl/KoehnHBCFBCSMZDBCH07}, we implement encoders and decoders as classes with the following (strongly simplified) interface:

  \begin{lstlisting}
  class Encoder {
    EncoderState build(Batch);
  };

  class Decoder {
    DecoderState startState(EncoderState[]);
    DecoderState step(DecoderState, Batch);
  };
  \end{lstlisting}

  A Bahdanau-style encoder-decoder model would implement the entire encoder inside \lstinline|Encoder::build| based on the content of the batch and place the resulting encoder context inside the \lstinline|EncoderState| object. 

  \lstinline|Decoder::startState| receives a list of \lstinline|EncoderState| (one in the case of the Bahdanau model, multiple for multi-source models, none for language models) and creates the initial \lstinline|DecoderState|. 

  The \lstinline|Decoder::step| function consumes the target part of a batch to produce the output logits of a model. The time dimension is either expanded by broadcasting of single tensors or by looping over the individual time-steps (for instance in the case of RNNs). 
  Loops and other control structures are just the standard built-in C++ operations. 
  The same function can then be used to expand over all given time steps at once during training and scoring or step-by-step during translation. 
  Current hypotheses state (e.g.~RNN vectors) and current logits are placed in the next \lstinline|DecoderState| object. 

  Decoder states are used mostly during translation to select the next set of translation hypotheses. Complex encoder-decoder models can derive from \lstinline|DecoderState| to implement non-standard selection behavior, for instance hard-attention models need to increase attention indices based on the top-scoring hypotheses.

  This framework makes it possible to combine different encoders and decoders (e.g.~RNN-based encoder with a Transformer decoder) and reduces implementation effort. In most cases it is enough to implement a single inference step in order to train, score and translate with a new model. 

  \subsection{Efficient Meta-algorithms}

  On top of the auto-diff engine and encoder-decoder framework we implemented many efficient meta-algorithms. These include multi-device (GPU or CPU) training, scoring and batched beam search, ensembling of heterogeneous models (e.g.~Deep RNN models and Transformer or language models), multi-node training and more. 

  \section{Case Studies}\label{case}
  In this section we will illustrate how we used the Marian toolkit to facilitate our own research across several NLP problems. 
  Each subsection is meant as a showcase for different components of the toolkit and demonstrates the maturity and flexibility of the toolkit. Unless stated otherwise, all mentioned features are included in the Marian toolkit.  

  \subsection{Improving over WMT2017 systems}

  \newcite{DBLP:conf/wmt/SennrichBCGHHBW17} proposed the highest scoring NMT system in terms of BLEU during the WMT 2017 shared task on English-German news translation \cite{DBLP:conf/wmt/2017}, trained with the Nematus toolkit \cite{sennrich-EtAl:2017:EACLDemo}. In this section, we demonstrate that we can replicate and slightly outperform these results with an identical model architecture implemented in Marian and improve on the recipe with a Transformer-style \cite{NIPS2017_7181} model. 

  \subsubsection{Deep Transition RNN Architecture}

  The model architecture in \newcite{DBLP:conf/wmt/SennrichBCGHHBW17} is a sequence-to-sequence model with single-layer RNNs in both, the encoder and decoder. The RNN in the encoder is bi-directional. Depth is achieved by building stacked GRU-blocks resulting in very tall RNN cells for every recurrent step (deep transitions). The encoder consists of four GRU-blocks per cell, the decoder of eight GRU-blocks with an attention mechanism placed between the first and second block. As in \newcite{DBLP:conf/wmt/SennrichBCGHHBW17}, embeddings size is 512, RNN state size is 1024. We use layer-normalization \cite{ba2016layer} and variational drop-out with $p=0.1$ \cite{gal2016theoretically} inside GRU-blocks and attention.

  \subsubsection{Transformer Architecture}
  We very closely follow the architecture described in \newcite{NIPS2017_7181} and their "base" model. 

  \subsubsection{Training Recipe}

  \begin{table}[t]
  \centering
  \begin{tabular}{lcc}\toprule
  System &	test2016 &	test2017 \\ \midrule
  UEdin WMT17 (single) & 33.9 & 27.5 \\ 
  +Ensemble of 4 & 35.1 & 28.3 \\ 
  +R2L Reranking & 36.2 &	28.3 \\ \midrule \midrule
  Deep RNN (single) & 34.3 & 27.7 \\ 
  +Ensemble of 4 & 35.3 & 28.2 \\ 
  +R2L Reranking & 35.9 & 28.7 \\ \midrule
  Transformer (single) &	35.6 &	28.8 \\
  +Ensemble of 4 &	36.4 &	29.4 \\ 
  +R2L Reranking &	36.8 &	29.5 \\ \bottomrule
  \end{tabular}
  \caption{BLEU results for our replication of the UEdin WMT17 system for the en-de news translation task.
  We reproduced most steps and replaced the deep RNN model with a Transformer model.}
  \label{wmt-bleu}
  \end{table}

  Modeled after the description\footnote{The entire recipe is available in form of multiple scripts at \url{https://github.com/marian-nmt/marian-examples}.} from \newcite{DBLP:conf/wmt/SennrichBCGHHBW17}, we perform the following steps:

  \begin{itemize}
  \item preprocessing of training data, tokenization, true-casing\footnote{Proprocessing was performed using scripts from Moses \cite{conf/acl/KoehnHBCFBCSMZDBCH07}.}, vocabulary reduction to 36,000 joint BPE subword units \cite{sennrich2016bpe} with a separate tool.\footnote{\url{https://github.com/rsennrich/subword-nmt}}
  \item training of a shallow model for back-translation on parallel WMT17 data;
  \item translation of 10M German monolingual news sentences to English; concatenation of artificial training corpus with original data (times two) to produce new training data;
  \item training of four left-to-right (L2R) deep models (either RNN-based or Transformer-based);
  \item training of four additional deep models with right-to-left (R2L) orientation; \footnote{R2L training, scoring or decoding does not require data processing, right-to-left inversion is built into Marian.}
  \item ensemble-decoding with four L2R models resulting in an n-best list of 12 hypotheses per input sentence;
  \item rescoring of n-best list with four R2L models, all model scores are weighted equally; 
  \item evaluation on newstest-2016 (validation set) and newstest-2017 with sacreBLEU.\footnote{\url{https://github.com/mjpost/sacreBLEU}}
  \end{itemize}

  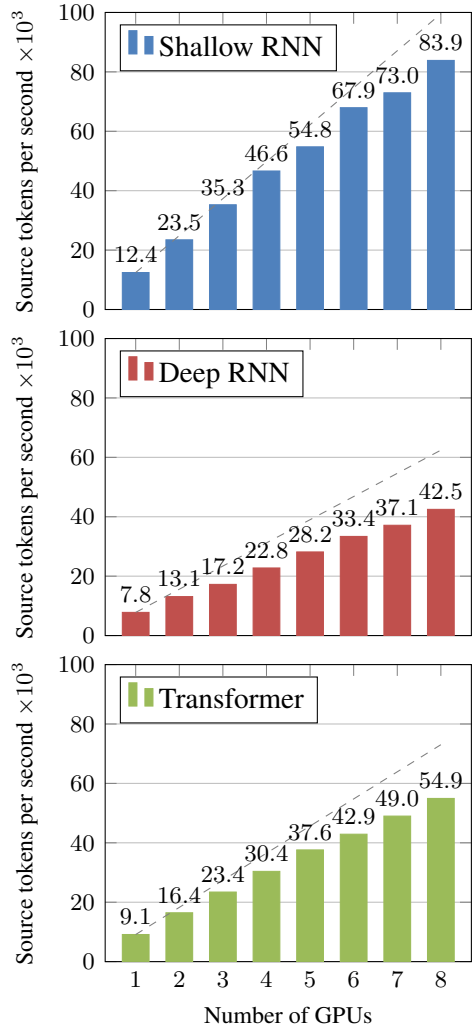
\begin{figure}[t]
  \centering
  \begin{tikzpicture}
  \begin{groupplot}[
      group style={group size= 1 by 3, vertical sep=1em,
       ylabels at=edge left,
       yticklabels at=edge left,
       xlabels at=edge bottom,
       xticklabels at=edge bottom,
       horizontal sep=5pt},
       ytick scale label code/.code={},
      small, width=0.4\textwidth,
      ymajorgrids,
      xlabel={Number of GPUs}, ylabel={Source tokens per second $\times 10^3$},
      ymin=0, ymax=100000,
      xtick={1 ,2 ,3, 4 ,5, 6, 7, 8},
      legend style={at={(0.04,0.96)},anchor=north west},
      nodes near coords style={font=\small, text=black,/pgf/number format/.cd,fixed zerofill,precision=1}]
  \nextgroupplot[scaled y ticks=base 10:-3]

  \addlegendimage{bblue, refstyle=barplot1}
  \addlegendentry{Shallow RNN}
  \addplot[ybar=0, ybar legend, bblue, fill, visualization depends on={rawy/1e3 \as \scaledy},
      nodes near coords={\pgfmathprintnumber{\scaledy}}] plot coordinates {
  (1, 12444) (2, 23458) (3,35262) (4, 46630) (5, 54765) (6, 67936) (7, 72953) (8, 83855)};
  \label{barplot1}

  \addplot[gray, dashed] plot coordinates {
  (1, 12444) (8, 99552)
  };

  \nextgroupplot[scaled y ticks=base 10:-3]
  \addlegendimage{rred, refstyle=barplot2}
  \addlegendentry{Deep RNN}
  \addplot+[ybar=0, ybar legend, rred, mark=none, fill, visualization depends on={rawy/1e3 \as \scaledy},	nodes near coords={\pgfmathprintnumber{\scaledy}}] plot coordinates {
  (1, 7802) (2, 13119) (3, 17230) (4, 22756) (5, 28154) (6, 33374) (7, 37134) (8, 42487)
  };
  \label{barplot2}

  \addplot[gray, dashed] plot coordinates {
  (1, 7802) (8, 62416)
  };

  \nextgroupplot[scaled y ticks=base 10:-3]
  \addlegendimage{ggreen, refstyle=barplot3}
  \addlegendentry{Transformer}
  \addplot+[ybar=0, ggreen, ybar legend, mark=none, fill, visualization depends on={rawy/1e3 \as \scaledy},
      nodes near coords={\pgfmathprintnumber{\scaledy}}] plot coordinates {
  (1, 9129) (2, 16445) (3, 23424) (4, 30381) (5, 37608)
  (6, 42897) (7, 48995) (8, 54938)
  };
  \label{barplot3}

  \addplot[gray, dashed] plot coordinates {
  (1, 9129) (8, 73032)
  };

  \end{groupplot}
  \end{tikzpicture}
  \caption{Training speed in thousands of source tokens per second for shallow RNN, deep RNN and Transformer model. Dashed line projects  linear scale-up based on single-GPU performance.}
  \label{wmt-speed-train}
  \end{figure}

  We train the deep models with synchronous Adam on 8 NVIDIA Titan X Pascal GPUs with 12GB RAM for 7 epochs each. The back-translation model is trained with asynchronous Adam on 8 GPUs. We do not specify a batch size as Marian adjusts the batch based on available memory to maximize speed and memory usage.  This guarantees that a chosen memory budget will not be exceeded during training.

  All models use tied embeddings between source, target and output embeddings \cite{press2017using}. Contrary to \newcite{DBLP:conf/wmt/SennrichBCGHHBW17} or \newcite{NIPS2017_7181}, we do not average checkpoints, but maintain a continuously updated exponentially averaged model over the entire training. Following \newcite{NIPS2017_7181}, the learning rate is set to 0.0003 and decayed as the inverse square root of the number of updates after 16,000 updates. When training the transformer model, a linearly growing learning rate is used during the first 16,000 iterations, starting with 0 until the base learning rate is reached. 

  \subsubsection{Performance and Results}

  \begin{table}[t]
  \centering
  \begin{tabular}{lccc}\toprule
  Model & 1 & 8 & 64 \\ \midrule
  Shallow RNN & 112.3 & 25.6 & 15.7\\
  Deep Transition RNN & 179.4 & 36.5 & 21.0\\
  Transformer & 362.7 & 98.5 & 71.3\\ \bottomrule
  \end{tabular}
  \caption{Translation time in seconds for newstest-2017 (3,004 sentences, 76,501 source BPE tokens) for different architectures and batch sizes.}
  \label{tab-trans}
  \end{table}

  \begin{filecontents*}{m-cgru_1.dat}
x,y,v
0,0,0.4225
1,0,0.1506
2,0,0.1094
3,0,0.0789
4,0,0.0517
5,0,0.0694
6,0,0.0057
7,0,0.0233
8,0,0.0035
9,0,0.0107
10,0,0.0031
11,0,0.0711

0,1,0.0062
1,1,0.6849
2,1,0.2096
3,1,0.0276
4,1,0.0099
5,1,0.0127
6,1,0.0010
7,1,0.0050
8,1,0.0039
9,1,0.0042
10,1,0.0038
11,1,0.0312

0,2,0.0002
1,2,0.0028
2,2,0.8925
3,2,0.0187
4,2,0.0288
5,2,0.0501
6,2,0.0015
7,2,0.0039
8,2,0.0001
9,2,0.0001
10,2,0.0001
11,2,0.0011

0,3,0.0004
1,3,0.0008
2,3,0.0044
3,3,0.8819
4,3,0.0323
5,3,0.0450
6,3,0.0280
7,3,0.0003
8,3,0.0004
9,3,0.0006
10,3,0.0013
11,3,0.0047

0,4,0.0001
1,4,0.0003
2,4,0.0004
3,4,0.0076
4,4,0.8411
5,4,0.1194
6,4,0.0253
7,4,0.0027
8,4,0.0002
9,4,0.0002
10,4,0.0001
11,4,0.0026

0,5,0.0002
1,5,0.0002
2,5,0.0004
3,5,0.0022
4,5,0.0260
5,5,0.6811
6,5,0.2359
7,5,0.0029
8,5,0.0003
9,5,0.0008
10,5,0.0006
11,5,0.0494

0,6,0.0000
1,6,0.0000
2,6,0.0001
3,6,0.0006
4,6,0.0045
5,6,0.0763
6,6,0.8985
7,6,0.0115
8,6,0.0009
9,6,0.0004
10,6,0.0003
11,6,0.0068

0,7,0.0007
1,7,0.0005
2,7,0.0012
3,7,0.0006
4,7,0.0050
5,7,0.0114
6,7,0.0088
7,7,0.9387
8,7,0.0093
9,7,0.0050
10,7,0.0017
11,7,0.0170

0,8,0.0000
1,8,0.0001
2,8,0.0001
3,8,0.0011
4,8,0.0010
5,8,0.0103
6,8,0.0695
7,8,0.0418
8,8,0.8518
9,8,0.0083
10,8,0.0008
11,8,0.0151

0,9,0.0004
1,9,0.0019
2,9,0.0032
3,9,0.0008
4,9,0.0101
5,9,0.0112
6,9,0.0056
7,9,0.1044
8,9,0.0646
9,9,0.6772
10,9,0.0143
11,9,0.1063

0,10,0.0002
1,10,0.0049
2,10,0.0026
3,10,0.0029
4,10,0.0077
5,10,0.0596
6,10,0.0088
7,10,0.0076
8,10,0.0479
9,10,0.0524
10,10,0.0251
11,10,0.7801

0,11,0.0007
1,11,0.0039
2,11,0.0020
3,11,0.0013
4,11,0.0046
5,11,0.0104
6,11,0.0042
7,11,0.0622
8,11,0.0190
9,11,0.3861
10,11,0.2773
11,11,0.2284

0,12,0.0007
1,12,0.0047
2,12,0.0087
3,12,0.0008
4,12,0.0022
5,12,0.0062
6,12,0.0040
7,12,0.0111
8,12,0.0221
9,12,0.1538
10,12,0.4162
11,12,0.3695

0,13,0.0008
1,13,0.0062
2,13,0.0027
3,13,0.0006
4,13,0.0025
5,13,0.0020
6,13,0.0017
7,13,0.0029
8,13,0.0062
9,13,0.0325
10,13,0.3545
11,13,0.5875

0,14,0.0021
1,14,0.0135
2,14,0.0096
3,14,0.0004
4,14,0.0141
5,14,0.0064
6,14,0.0044
7,14,0.0251
8,14,0.0064
9,14,0.0704
10,14,0.1113
11,14,0.7362
\end{filecontents*}
\DTLloaddb[headers={x,y,v}, keys={x,y,v}]{dataA}{m-cgru_1.dat}
  \begin{filecontents*}{m-cgru_2.dat}
x,y,v
0,0,0.9432
1,0,0.0176
2,0,0.0035
3,0,0.0026
4,0,0.0083
5,0,0.0031
6,0,0.0005
7,0,0.0006
8,0,0.0001
9,0,0.0204

0,1,0.3566
1,1,0.3236
2,1,0.0146
3,1,0.0219
4,1,0.0527
5,1,0.0161
6,1,0.0021
7,1,0.0224
8,1,0.0233
9,1,0.1667

0,2,0.0060
1,2,0.7571
2,2,0.0371
3,2,0.0199
4,2,0.0680
5,2,0.0156
6,2,0.0005
7,2,0.0200
8,2,0.0169
9,2,0.0589

0,3,0.0041
1,3,0.0624
2,3,0.8162
3,3,0.0450
4,3,0.0063
5,3,0.0007
6,3,0.0006
7,3,0.0018
8,3,0.0034
9,3,0.0595

0,4,0.0206
1,4,0.0792
2,4,0.1509
3,4,0.1302
4,4,0.1701
5,4,0.0253
6,4,0.0108
7,4,0.0198
8,4,0.0266
9,4,0.3666

0,5,0.0038
1,5,0.0790
2,5,0.0772
3,5,0.1484
4,5,0.1729
5,5,0.0075
6,5,0.0127
7,5,0.0034
8,5,0.0178
9,5,0.4773

0,6,0.0014
1,6,0.0051
2,6,0.0331
3,6,0.0227
4,6,0.1786
5,6,0.0142
6,6,0.0029
7,6,0.0096
8,6,0.0061
9,6,0.7263

0,7,0.0084
1,7,0.0225
2,7,0.0123
3,7,0.0597
4,7,0.4751
5,7,0.2215
6,7,0.0196
7,7,0.0090
8,7,0.0410
9,7,0.1310

0,8,0.0009
1,8,0.0008
2,8,0.0055
3,8,0.0432
4,8,0.0228
5,8,0.0705
6,8,0.4589
7,8,0.2773
8,8,0.0114
9,8,0.1086

0,9,0.0336
1,9,0.0316
2,9,0.0032
3,9,0.0723
4,9,0.0437
5,9,0.0095
6,9,0.4511
7,9,0.0223
8,9,0.1574
9,9,0.1753

0,10,0.0003
1,10,0.0866
2,10,0.0106
3,10,0.0068
4,10,0.0062
5,10,0.0031
6,10,0.7295
7,10,0.0059
8,10,0.0151
9,10,0.1358

0,11,0.0155
1,11,0.2183
2,11,0.0183
3,11,0.0454
4,11,0.0154
5,11,0.0040
6,11,0.0556
7,11,0.0084
8,11,0.2917
9,11,0.3275

0,12,0.0715
1,12,0.1281
2,12,0.0329
3,12,0.0163
4,12,0.0126
5,12,0.0075
6,12,0.0046
7,12,0.0032
8,12,0.1357
9,12,0.5875

0,13,0.0600
1,13,0.1122
2,13,0.0732
3,13,0.0197
4,13,0.0081
5,13,0.0082
6,13,0.0133
7,13,0.0033
8,13,0.1046
9,13,0.5974

0,14,0.0718
1,14,0.0610
2,14,0.0328
3,14,0.0405
4,14,0.0152
5,14,0.0046
6,14,0.0072
7,14,0.0089
8,14,0.1336
9,14,0.6245
\end{filecontents*}
\DTLloaddb[headers={x,y,v}, keys={x,y,v}]{dataB}{m-cgru_2.dat}

  \begin{figure*}[t]
  \centering
  \def\drawalignmentA{}%
  \DTLforeach*{dataA}{\x=x,\y=y,\v=v}{
      \appto\drawalignmentA{\fill}
      \eappto\drawalignmentA{ [black, fill=bblue, fill opacity=\v, thick] (axis cs:\x,\y) rectangle (axis cs:\x+1,\y+1);}%
  }
  \def\drawalignmentB{}%
  \DTLforeach*{dataB}{\x=x,\y=y,\v=v}{
      \appto\drawalignmentB{\fill}
      \eappto\drawalignmentB{ [black, fill=bblue, fill opacity=\v, thick] (axis cs:\x,\y) rectangle (axis cs:\x+1,\y+1);}%
  }
  \begin{tikzpicture}
  \begin{groupplot}[
    group style={
       group size=2 by 1,
       horizontal sep=10pt,
       group name=G},
    x=11pt, y=11pt,
    xticklabel style={xshift=4pt,rotate=45,anchor=west,yshift=2pt},
    yticklabel style={yshift=-6pt}]

    \nextgroupplot[
     y dir=reverse,
     axis background/.style={draw=black, line width=.2pt},
     axis x line*=top,
    xmin=0, xmax=12,
    ymin=0, ymax=15,
    xtick={0,1,2,3,4,5,6,7,8,9,10,11},
    ytick={0,1,2,3,4,5,6,7,8,9,10,11,12,13,14},
    xticklabels={W\"{a}hlen, Sie, einen, Tastatur-, be-, fehl-, ssatz, im, Men\"{u}, festlegen, ., EOS},
    yticklabels={W\"{a}hlen, Sie, einen, Tastatur-, be-, fehl-, ssatz, im, Men\"{u}, ``, Satz, '', aus, ., EOS},
    tick style={draw=none},
    tick label style={font=\footnotesize},
    xlabel absolute,
    xlabel={$\mathrm{mt}$},
    grid=both,
    xlabel style={yshift=2ex}] 
  \drawalignmentA
  \draw[rred,thick] (axis cs:0,10) rectangle (axis cs:13,11);
   \nextgroupplot[
    y dir=reverse,
    axis background/.style={draw=black, line width=.2pt},
    axis x line*=top,
    xmin=0, xmax=10,
    ymin=0, ymax=15,
    xtick={0,1,2,3,4,5,6,7,8,9,10},
    ytick={0,1,2,3,4,5,6,7,8,9,10,11,12,13,14},
    yticklabels={W\"{a}hlen, Sie, einen, Tastatur-, be-, fehl-, ssatz, im, Men\"{u}, ``, Satz, '', aus, ., EOS},
    yticklabel pos=right,
    xticklabels={Select, a, shortcut, set, in, the, Set, menu, ., EOS},
    tick style={draw=none},
    grid=both,
    tick label style={font=\footnotesize},
    xlabel={$\mathrm{src}$},
    xlabel absolute,
    xlabel style={yshift=2ex},
  ] \drawalignmentB
  \draw[rred,thick] (axis cs:0,10) rectangle (axis cs:10,11);
  \draw[rred,thick] (axis cs:6,0) rectangle (axis cs:7,15);
  \end{groupplot}
  \end{tikzpicture}
  \vspace{-2mm}
  \caption{Example for error recovery based on dual attention. The missing word ``Satz'' could only be recovered based on the original source (marked in red) as it was dropped in the raw MT output.}
  \label{ape-rec}
  \end{figure*}
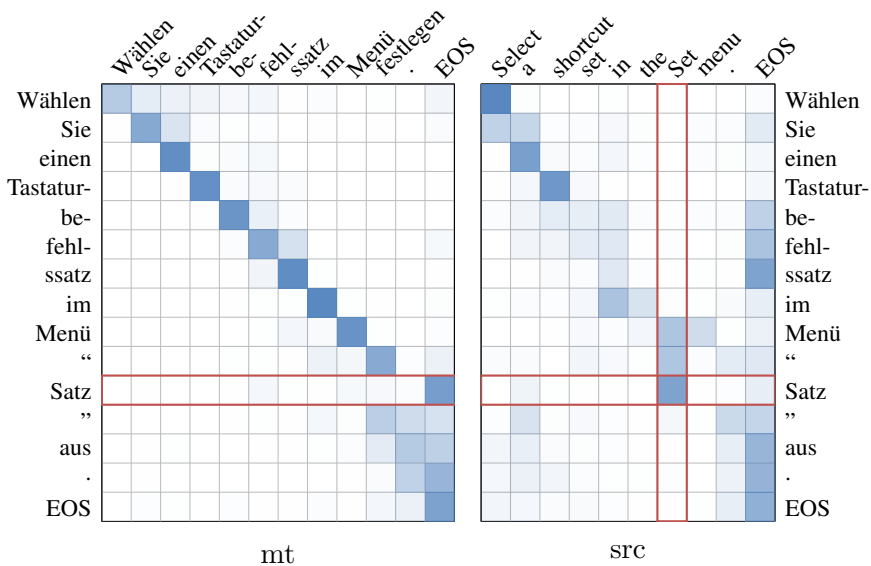

  \paragraph{Quality.} In terms of BLEU (Table~\ref{wmt-bleu}), we match  the original Nematus models from \newcite{DBLP:conf/wmt/SennrichBCGHHBW17}. 
  Replacing the deep-transition RNN model with the transformer model results in a significant BLEU improvement of 1.2 BLEU on the WMT2017 test set. 

  \paragraph{Training speed.}

  In Figure~\ref{wmt-speed-train} we demonstrate the training speed as thousands of source tokens per second for the models trained in this recipe. All model types benefit from using more GPUs. Scaling is not linear (dashed lines), but close. The tokens-per-second rate (w/s) for Nematus on the same data on a single GPU is about 2800 w/s for the shallow model. Nematus does not have multi-GPU training. Marian achieves about 4 times faster training on a single GPU and about 30 times faster training on 8 GPUs for identical models.

  \paragraph{Translation speed.}

  The back-translation of 10M sentences with a shallow model takes about four hours on 8 GPUs at a speed of about 15,850 source tokens per second at a beam-size of 5 and a batch size of 64. Batches of sentences are translated in parallel on multiple GPUs. 

  In Table~\ref{tab-trans} we report the total number of seconds to translate newstest-2017 (3,004 sentences, 76,501 source BPE tokens) on a single GPU for different batch sizes. We omit model load time (usually below 10s). Beam size is 5.

  \subsection{State-of-the-art in Neural Automatic Post-Editing}
  In our submission to the Automatic Post-Editing shared task at WMT-2017 \cite{bojar-EtAl:2017:WMT1} and follow-up work \cite{junczysdowmunt-grundkiewicz:2017:WMT, I17-1013}, 
  we explore multiple neural architectures adapted for the task of automatic post-editing of machine translation output as implementations in Marian. We focus on neural end-to-end models that combine both inputs $mt$ (raw MT output) and $src$ (source language input) in a single neural architecture, modeling $\{mt,src\}\rightarrow pe$ directly, where $pe$ is post-edited corrected output.

  These models are based on multi-source neural translation models introduced by \newcite{DBLP:journals/corr/ZophK16}. Furthermore, we investigate the effect of hard-attention models or neural transductors \cite{aharoni2016sequence} which seem to be well-suited for monolingual tasks, as well as combinations of both ideas. 
  Dual-attention models that are combined with hard attention remain competitive despite applying fewer changes to the input.

  The encoder-decoder framework described in section~\ref{encdec}, allowed to integrate dual encoders and hard-attention without changes to beam-search or ensembling mechanisms.
  The dual-attention mechanism over two encoders allowed to recover missing words that would not be recognized based on raw MT output alone, see Figure~\ref{ape-rec}.

  Our final system for the APE shared task scored second-best according to automatic metrics and best based on human evaluation.



  \subsection{State-of-the-art in Neural Grammatical Error Correction}

  In \newcite{mjdrggec}, we use Marian for research on transferring methods from low-resource NMT on the ground of automatic grammatical error correction (GEC).
  Previously, neural methods in GEC did not reach state-of-the-art results compared to phrase-based SMT baselines. We successfully adapt several low-resource MT methods for GEC.

  We propose a set of model-independent methods for neural GEC that can be easily applied in most GEC settings.
  The combined effects of these methods result in better than state-of-the-art neural GEC models that outperform previously best neural GEC systems by more than 8\% M$^2$ on the CoNLL-2014 benchmark and more than 4.5\% on the JFLEG test set. Non-neural state-of-the-art systems are matched on the CoNLL-2014 benchmark and outperformed by 2\% on JFLEG.

  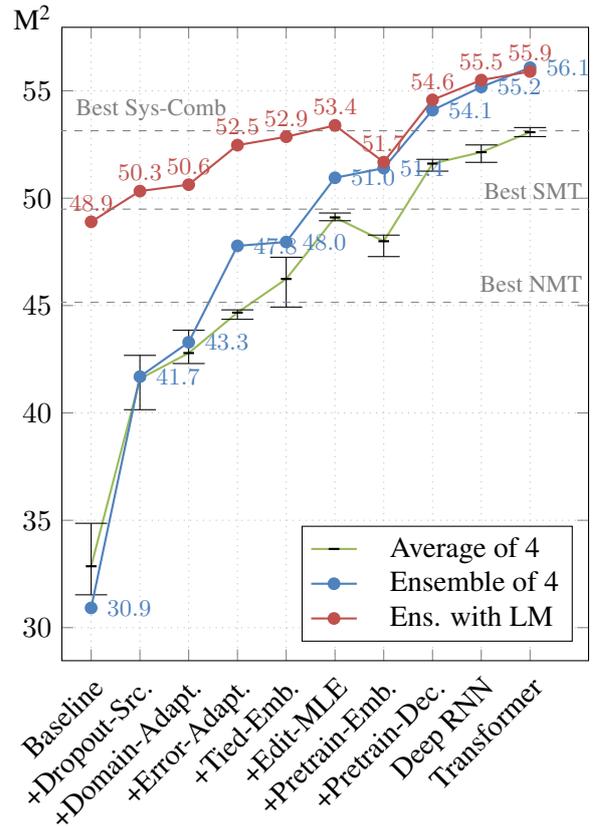
\begin{figure}[t]

  \begin{tikzpicture}
  \begin{axis}[
  ylabel=M\textsuperscript{2},
  every axis y label/.style={
       at={(ticklabel* cs:0.98)},
       anchor=south east, align=right
  },
  height=0.34\textheight,
  width=0.9\linewidth,
  scale only axis,
  enlarge y limits,
  ymajorgrids, xmajorgrids,
  major grid style={dotted},
  xtick={1,2,3,4,5,6,7,8,9,10},
  legend cell align=left,
  legend style={column sep=10pt},
  every node near coord/.append style={anchor=south, font=\small, xshift=0pt, /pgf/number format/.cd, fixed, fixed zerofill, precision=1, /tikz/.cd},
  xticklabels={Baseline,+Dropout-Src.,+Domain-Adapt.,+Error-Adapt.,
  +Tied-Emb.,+Edit-MLE,+Pretrain-Emb.,+Pretrain-Dec.,Deep RNN,Transformer},
  ymax=55.5,
  xmin=0.4,
  xmax=11.2,
  legend pos=south east,
  xticklabel style={align=right, rotate=45, anchor=north east},
  ]
  \addplot+[solid, thick, ggreen, mark=-, 
  mark options={solid, fill=black, black},
  error bars/.cd,y dir=both,y explicit, error bar style={solid, black},
      error mark options={
        rotate=90,
        mark size=6pt
      }]
  table[x index=0,
        y expr={(\thisrow{m1}+\thisrow{m2}+\thisrow{m3}+\thisrow{m4})/4},
        y error plus expr={max(\thisrow{m1},\thisrow{m2},\thisrow{m3},\thisrow{m4})-(\thisrow{m1}+\thisrow{m2}+\thisrow{m3}+\thisrow{m4})/4},
        y error minus expr={-min(\thisrow{m1},\thisrow{m2},\thisrow{m3},\thisrow{m4})+(\thisrow{m1}+\thisrow{m2}+\thisrow{m3}+\thisrow{m4})/4},
        ] {test2014};

  \addplot+[nodes near coords,
  every node near coord/.append style={anchor=west, xshift=2pt},
  solid, thick, bblue, mark=*,mark options={solid, bblue, fill=bblue}]
  table[x index=0, y=ens] {test2014};

  \addplot+[solid, thick, rred, nodes near coords, mark=*, mark options={solid,fill=rred,rred}]
  table[x index=0, y=lm] {test2014};

   \draw[dashed, gray] (axis cs:0,53.14) -- (axis cs:12,53.14)
   node[below, anchor=south west, pos=0.04] {\small Best Sys-Comb};
   \draw[dashed, gray] (axis cs:0,49.49) -- (axis cs:12,49.49)
   node[below, anchor=south east, pos=0.94] {\small Best SMT};

   \draw[dashed, gray] (axis cs:0,45.15) -- (axis cs:12,45.15)
   node[anchor=south east, pos=0.94] {\small Best NMT};

  \legend{Average of 4, Ensemble of 4, Ens. with LM}
  \end{axis}
  \end{tikzpicture}
  \vspace{-10mm}
  \caption{Comparison on the CoNLL-2014 test set for investigated methods.}\label{fig.all}
  \end{figure}

  Figure~\ref{fig.all} illustrates these results on the CoNLL-2014 test set. 
  To produce this graph, 40 GEC models (four per entry) and 24 language models (one per GEC model with pre-training) have been trained.  The language models follow the decoder architecture and can be used for transfer learning, weighted decode-time ensembling and re-ranking. This also includes a Transformer-style language model with self-attention layers. 

  Proposed methods include extensions to Marian, such as source-side noise, a GEC-specific weighted training-objective, usage of pre-trained embeddings, transfer learning with pre-trained language models, decode-time ensembling of independently trained GEC models and language models, and various deep architectures.

  \section{Future Work and Conclusions}
  We introduced Marian, a self-contained neural machine translation toolkit written in C++ with focus on efficiency and research. Future work on Marian's back-end will look at faster CPU-bound computation, auto-batching mechanisms and automatic kernel fusion. On the front-end side we hope to keep up with future state-of-the-art models.

  \section*{Acknowledgments}

  {\small The development of Marian received funding from the European Union's Horizon 2020 Research and Innovation Programme under grant agreements 688139 (SUMMA; 2016-2019), 645487 (Modern MT; 2015-2017), 644333 (TraMOOC; 2015-2017), 644402 (HimL; 2015-2017), the Amazon Academic Research Awards program, and the World Intellectual Property Organization. The CPU back-end was contributed by Intel under a partnership with the Alan Turing Institute. 

   This research is based upon work supported in part by the Office of the Director of National Intelligence (ODNI), Intelligence Advanced Research Projects Activity (IARPA), via contract \#FA8650-17-C-9117. The views and conclusions contained herein are those of the authors and should not be interpreted as necessarily representing the official policies, either expressed or implied, of ODNI, IARPA, or the U.S. Government. The U.S. Government is authorized to reproduce and distribute reprints for governmental purposes notwithstanding any copyright annotation therein.\par}

  \bibliography{marian}
  \bibliographystyle{acl_natbib}
  \end{document}